\documentclass[letterpaper]{article} 
\usepackage{times}  
\usepackage{helvet}  
\usepackage{courier}  
\usepackage[hyphens]{url}  
\usepackage{graphicx} 
\usepackage{natbib}  
\usepackage{caption} 
\usepackage{caption3} 
\usepackage[symbol]{footmisc}
\usepackage{subfig}
\usepackage{graphicx}     
\usepackage{enumitem}
\usepackage{caption} 
\usepackage{amsmath}
\usepackage{amssymb}
\usepackage{tikz} 
\usetikzlibrary{automata,positioning,calc, arrows}
\usepackage{booktabs}
\usepackage{numprint}
\usepackage{siunitx}
\npdecimalsign{.}
\nprounddigits{4}
\usepackage{algorithm}
\usepackage{algorithmic}
\usepackage{xcolor}
\usepackage{newfloat}
\usepackage{listings}
\usepackage[affil-it]{authblk}
\floatstyle{ruled}
\newfloat{listing}{tb}{lst}{}
\floatname{listing}{Listing}
\title{Leveraging Interpretability in the Transformer to Automate the Proactive Scaling of Cloud Resources}

\author[1]{Amadou Ba}
\author[2]{Pavithra Harsha}
\author[2]{  Chitra Subramanian}
\affil[1]{IBM Research Europe, Dublin}
\affil[2]{IBM T. J Watson Research Center, Yorktown Heights, NY 10570s}

\date{}

\begin{document}

\maketitle

\begin{abstract}
Modern web services adopt cloud-native principles to leverage the advantages of microservices. To consistently guarantee high Quality of Service (QoS) according to Service Level Agreements (SLAs), ensure satisfactory user experiences, and minimize operational costs, each microservice must be provisioned with the right amount of resources. However, accurately provisioning microservices with adequate resources is complex and depends on many factors, including workload intensity and the complex interconnections between microservices. To address this challenge, we develop a model that captures the relationship between an end-to-end latency, requests at the front-end level, and resource utilization. We then use the developed model to predict the end-to-end latency. Our solution leverages the Temporal Fusion Transformer (TFT), an attention-based architecture equipped with interpretability features. When the prediction results indicate SLA non-compliance, we use the feature importance provided by the TFT as covariates in Kernel Ridge Regression (KRR), with the response variable being the desired latency, to learn the parameters associated with the feature importance. These learned parameters reflect the adjustments required to the features to ensure SLA compliance. We demonstrate the merit of our approach with a microservice-based application and provide a roadmap to deployment.
\end{abstract}

\section{Introduction}
One of the primary motivations driving application developers toward cloud systems is the possible access to large-scale infrastructure and automation platforms, guaranteeing scalability, flexibility, and cost-effectiveness, among a myriad of other advantages. To ensure high Quality of Service (QoS) for application developers, cloud providers attempt to strike an optimal balance between resource usage and QoS. However, provisioning resources to meet application performance while minimizing wastage and costs presents a challenge \cite{lee}. This challenge is further amplified by the rise of cloud-native tools that promote microservice architectures, which seek to ensure end-to-end QoS in production environments \cite{roy}. Microservice architectures are characterized by complex request execution paths that traverse multiple microservices. The latency from one microservice impacts the latency of downstream microservices, thereby affecting the end-to-end latency of the entire application trace. To prevent high latency in microservices-based applications and ensure QoS, autoscaling approaches have been developed. These approaches are either reactive or proactive \cite{calzarossa, roy}. Reactive autoscaling solutions make scaling decisions by analyzing current system metrics, such as CPU utilization and memory usage. The autoscaling mechanism is triggered when these metrics show abnormalities. However, reactive autoscaling is limited due to its inability to act beforehand to prevent QoS degradation. This limitation led to the development of proactive autoscaling approaches, which are based on either predicted workload \cite{wang, xue} or predicted end-to-end latency \citep{joy, haytham, luo}. However, existing prediction approaches for autoscaling are not interpretable. Our work lies in this line of research, where we propose a new approach aiming to achieve interpretable prediction of end-to-end latency for the implementation of informed, fine-grained autoscaling mechanisms. This allows us to determine and scale specific microservices experiencing volatile demand, rather than scaling the entire application. To this end, we make the following contributions in this paper.
\textbf{(1)} We develop a model that captures the relationship between an end-to-end latency, requests at the front-end level and resource utilization. Then, \textbf{(2)} we use the developed model to predict the end-to-end latency. Our solution leverages the Temporal Fusion Transformer (TFT) \cite{lim}. The TFT utilizes recurrent layers for local processing to learn temporal relationships at different scales and is equipped with interpretable self-attention layers for capturing long-term dependencies. This dual-layered structure enables our approach to simultaneously provide accurate predictions and interpretable results, offering us insights across the entire execution path and paving the way for advanced resource provisioning. \textbf{(3)} Whenever the prediction results lead to SLA non–compliance, we use the feature importance provided by the TFT as covariates in a Kernel Ridge Regression (KRR), with the response variable being the desired latency, to learn the parameters associated with the feature importance. The learned parameters allow us to perform autoscaling when they are associated to resource usage. \textbf{(4)} We demonstrate the viability of our approach in a practical setting characterized by a cloud-native application.
To the best of our knowledge, this is the first paper to develop a mechanism for interpretable prediction of end-to-end latency for microservices-based applications, and to use the interpretability results as a basis for autoscaling cloud resources.
\section{Related work}
\label{sec:related work}
To proactively provision resources to microservices, machine learning (ML) and deep learning (DL) approaches are increasingly being utilized \citep{joy, haytham, luo}. These approaches aim to efficiently and optimally adjust resource allocation \citep{aslanpour, zhou}. To enhance the relevance of the  ML and DL models used for autoscaling, domain knowledge represented by causal mechanisms is gradually being considered in these models. Their objective is to capture the interrelations between the components of the microservices \citep{yanqi, chow, liang}. Generally, these causal mechanisms are represented by a graph, and Graph Neural Networks are employed to model the causal relations \citep{tam, park, wang}. However, their operationalization presupposes perfect knowledge and a complete representation of all interconnections characterizing the topology of microservices. Furthermore, these approaches do not provide interpretable predictions of end-to-end latency for proactive autoscaling. This is where the contributions of our work lie, where we provide interpretable and actionable predictions of end-to-end latency.

\section{Approach to proactive autoscaling}
\label{model}
\begin{figure}[t!]
\centering
\includegraphics[width=0.99\columnwidth]{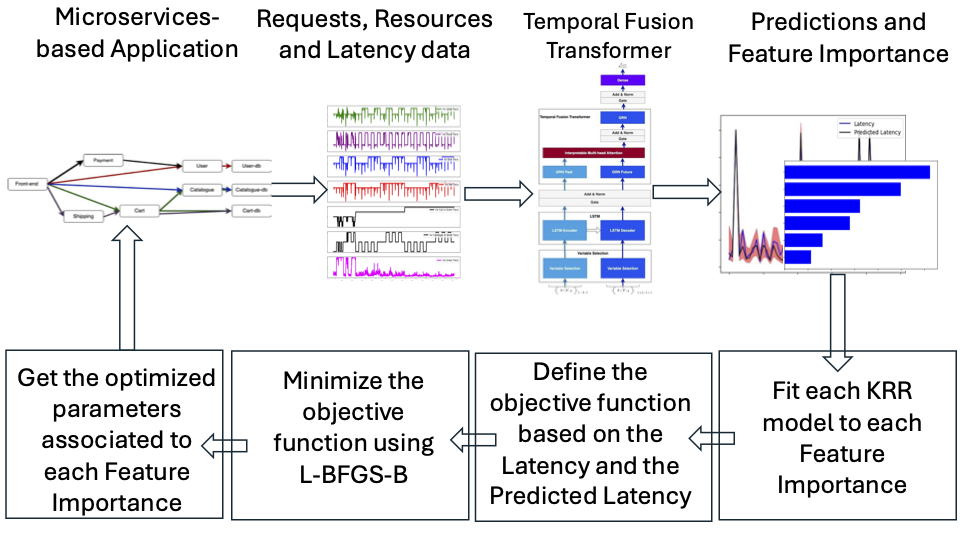}
\caption{Building blocks of the proposed approach. (1) Deploying the microservices and acquiring the data. (2) Building the predictive models using Temporal Fusion Transformer. (3) Using the statistical feature importance values as new features for the KRR and building the new predictive models by fitting each KRR model to each feature importance. (4) Defining and minimizing the objective function based on the actual latency and the predicted latency. (5) Using the estimated parameters associated with each feature importance to perform autoscaling.}
\label{fig:approach}
\end{figure}

Figure \ref{fig:approach} presents our approach to proactive autoscaling of cloud resources. The approach starts with using the TFT to predict an end-to-end latency, then it considers the statistical significance of the feature importance provided by the multi-head attention of the TFT to fit KRR. To estimate the parameters associated with each feature importance, we use an optimization algorithm from the class of quasi-Newton methods that approximates the Broyden–Fletcher–Goldfarb–Shanno algorithm. The estimated parameters are then used to perform autoscaling.

\subsection{Data representation for latency prediction}

We categorize the resources used in our approach into two types, \textit{vertical} and \textit{horizontal}. Vertical resources include those used in the container host, such as CPU and memory. Horizontal resources refer to the number of pod replicas used. Let's consider by $y_{t,m} \in \mathbb{R}^{N \times M}$ 
as the latency at time $t$ at the front-end level of an application, where $m=1, \cdots, M$ represents a trace ID and $t=0, \cdots, N$ represents the time instant. The trace ID is used to track the flow of a single call as it traverses through various microservices. We consider that the inputs data at the trace level are given by $\mathcal{X} = \left\{x_{t,1}, \cdots, x_{t,M}\right\} $, where $x_{t,m} \in \mathbb{R}^{N \times M}$ are the features associated with the calls at the front-end level. The features at the microservices level are given by $\mathcal{X}^{\prime} = \left\{X_{t,1}^{L_{1}}, \cdots, X_{t,P}^{L_{P}}\right\}$, where $X_{t,p}^{L_{p}} \in \mathbb{R}^{N\times L_{P}}$, and $L_{p}$, $p =1, \cdots, P$, is the number of features $L$ at the microservice $p$. Our objective is to learn the function that maps an end-to-end latency to the calls at the front-end level and the features at the microservices level associated with the end-to-end latency. To this end, we use the TFT because of its advanced modeling capability and interpretability features.

\subsection{Temporal Fusion Transformer for the prediction}
\label{TFT}

We adopt an interpretable AI method based on the Transformer \cite{vaswani} for the interpretable prediction of end-to-end latency and the determination of the influential features. For this purpose, we use the Temporal Fusion Transformer (TFT), where we adopt certain modifications to suit our application. For example, we ignore the static covariates. The TFT is an AI model designed for time series prediction. It integrates the Transformer architecture with Temporal Fusion mechanisms to capture temporal patterns in sequential data. The TFT is composed of the multi–head attention mechanism from the Transformer with Recurrent Neural Networks (RNNs). Three main building blocks are present in the TFT, the variable selection networks, the Long Short-Term Memory (LSTM) encoder-decoder, and the interpretable multi-head attention. The encoder receives the data required for training and the decoder provides the predictions. The interpretability of the approach is given by the multi-head attention. The input layer to our TFT architecture is composed of the features $\mathcal{X}$ and $\mathcal{X}^{\prime}$ and the output layer combines the processed inputs with learned parameters to produce the quantile prediction of $y_{t, m}$. For the training phase, the inputs include past features $\left\{\mathcal{X} \cup \mathcal{X}^{\prime}, y\right\}_{t-k:t}$, characterized by the calls at the front-end level and the vertical and/or horizontal resources at the microservices level, along with the target variable, which is the end-to-end latency at the front-end level. These past features and the response variable are processed by the variable selection network before being passed to the LSTM encoder.
The output of the LSTM encoder is then fed to the Gated Recurrent Networks (GRN) after applying the gating mechanism, addition, and normalization operations. Subsequently, the output from the GRN, which receives the past information, is passed to the interpretable multi-head attention. For the scoring phase, the TFT architecture receives as input the future known features,  $\left\{\mathcal{X} \cup \mathcal{X}^{\prime}\right\}_{t+1:t+\tau}$, and after applying the variable selection, these features are fed to the LSTM decoder and their output to another GRN for processing, after gating, normalization and addition operations are applied. They produce the quantile prediction of the end-to-end latency.

\subsection{Kernel Ridge Regression for parametric estimation}
\label{Section:KRR}
Besides the quantile predictions of the end-to-end latency produced by the TFT, the multi-head attention mechanism of the TFT provides interpretability associated with these predictions. These interpretations are presented in the form of feature importance scores. The feature importance represents scores associated with the input features used in the TFT, based on their contributions to the prediction. Each score associated with a feature becomes a new sample for the KRR. The objective is to determine how much the features need to be readjusted after an SLA violation. Whenever an SLA violation occurs, we determine the percentage of violation in latency and define a desired latency by subtracting a factor from the predicted latency, making the new latency the desirable latency. We then learn the parameters linking the new features—comprised of scores from the multi-head attention interpretation—with the desired latency. These learned parameters represent autoscaling factors when they relate to actionable features such as the number of pods, CPU, memory, and so on.
We create $K$ KRR models, each corresponding to a new feature derived from the feature importance scores. Each KRR is trained on one of the new features, with the target variable being the newly defined latency. This step involves learning a function that maps each new feature to the desired latency. At this stage, we define a function that takes parameters $\theta_{0}, \theta_{1}, \cdots, \theta_{K}$, where $K$ is the number of features, and computes predictions by combining the outputs of the $K$ KRR models. Each KRR model predicts a nonlinear transformation of its corresponding feature, and the predictions are weighted and summed according to the parameters.

\subsection{The autoscaling mechanism}
\label{KRR}
We define the objective function to compute the squared difference between the combined model predictions of the desired end-to-end latency and the actual end-to-end latency. We then minimize this function to find the best parameters. To determine the parameters $\theta_{0}, \theta_{1}, \cdots, \theta_{K}$, we use the Limited-memory Broyden-Fletcher-Goldfarb-Shanno with Box constraints (L-BFGS-B) algorithm, which is an algorithm widely used for solving optimization problems with constraints on the variables. The advantage of using Box constraints in the optimization approach—corresponding to specifying bounds on the variables—lies in its ability to control how the autoscaling of cloud resources is performed.

\begin{figure}[ht]
\centering
\includegraphics[width=0.85\textwidth]{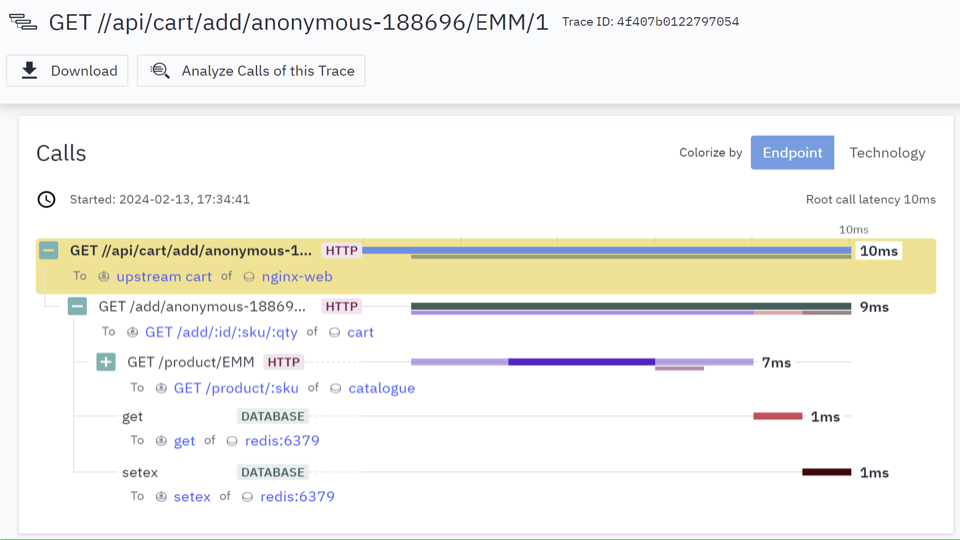}
\caption{Example of traces execution and their duration.}
\label{fig:trace}
\end{figure}
\section{Experiments}
\label{experiments}
\begin{figure}[h]
\centering
\includegraphics[width=0.85\textwidth]{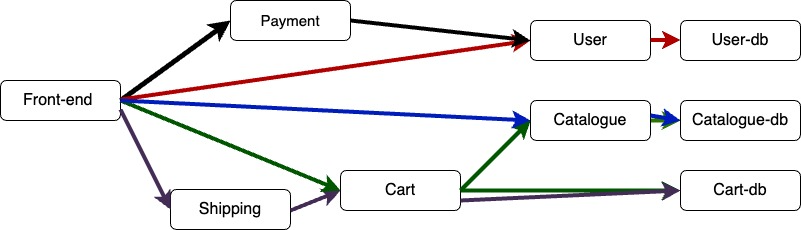}
\caption{Call graph using Robot Shop.}
\label{fig:microservice}
\end{figure}

Our use case focuses on Robot Shop, an e-commerce website that provides a comprehensive environment and functionalities. Some of the components present are the catalog, cart, payments, and shipping. Each of these components within the Robot Shop is represented by a distinct microservice. This architectural approach demonstrates the practical implementation of microservices and provides a robust platform for testing various resource provisioning mechanisms specific to a microservice environment. Robot Shop illustrates the advantages and intricacies of a microservices-based application. The microservices in Robot Shop are built using a variety of tools and technologies, reflecting a polyglot programming environment. This includes services developed in NodeJS, Java, Python, Golang, and PHP. The application leverages databases and messaging systems like MongoDB, Redis, MySQL, and RabbitMQ. Additionally, web server technologies like Nginx and front-end frameworks such as AngularJS are utilized. This diverse technological stack showcases how different programming languages and frameworks can be integrated to create a cohesive and functional application. Each microservice is encapsulated in a Docker container, for deployment and management by ensuring consistent runtime environments. These Dockerized services communicate with each other through REST APIs and messaging queues. The design emphasizes scalability and resilience, ensuring that the application can handle varying loads. In our use case, Robot Shop is deployed on IBM Cloud. The deployment is orchestrated with Kubernetes, which automates the deployment, scaling, and management of the pod replicas. Figure \ref{fig:trace} presents an example of traces execution and their duration, whereas Figure \ref{fig:microservice} presents a call graph of the Robot Shop that shows multiple execution paths as a result of concurrent client calls. In this example, we observe 5 traces, each following a specific execution path:

\begin{itemize}
    \item \textsf{Purple trace}: \texttt{Front-end} $\boldsymbol{\rightarrow}$ \texttt{Shipping} $\boldsymbol{\rightarrow}$ \texttt{Cart} $\boldsymbol{\rightarrow}$ \texttt{Cart-db}
       \item \textsf{Green trace}: \texttt{Front-end} $\boldsymbol{\rightarrow}$ \texttt{Cart} $\boldsymbol{\rightarrow}$ \texttt{Cart-db}
\item \textsf{Green trace}:  \texttt{Cart} $\boldsymbol{\rightarrow}$ \texttt{Catalogue}
$\boldsymbol{\rightarrow}$ \texttt{Catalogue-db}
\item \textsf{Blue trace}: \texttt{Front-end} $\boldsymbol{\rightarrow}$ \texttt{Catalogue}  $\boldsymbol{\rightarrow}$ \texttt{Catalogue-db}
\item \textsf{Red trace}: \texttt{Front-end} $\boldsymbol{\rightarrow}$ \texttt{User} $\boldsymbol{\rightarrow}$ \texttt{User-db}
\item \textsf{Black trace}: \texttt{Front-end} $\boldsymbol{\rightarrow}$ \texttt{Payment} $\boldsymbol{\rightarrow}$ \texttt{User} $\boldsymbol{\rightarrow}$ \texttt{User-db}
\end{itemize}

These traces show that the requests are characterized by call paths through service dependency graphs, which capture communication-based dependencies between microservice instances. The call paths illustrate how requests flow among microservices by following parent-child relationship chains. This demonstrates that service dependency graphs are important tools for discerning the complex interplay between services.
We consider this use case to demonstrate the performance of our proposed approach, which involves predicting an end-to-end latency at the trace level. For illustration purposes, we focus our experiments on the green and purple traces. 

\subsection{Selected end-to-end latency}

We consider the latency p95 as our response variable. The latency p95 (95th percentile) represents the latency value below which 95\% of the measured latency values fall. Compared to the latency p99, it provides a broader view of the latency distribution. Generally, latency p95 is employed to understand the performance of a system and is usually less impacted by extreme outliers compared to p99. If the latency p95 for the Robot Shop is 100 milliseconds, it means that 95\% of responses are received in 100 milliseconds or less, while 5\% of responses may experience longer latencies. As mentioned earlier, we developed our TFT model to take as input the calls at the front-end level across all the traces, in addition to the number of pods at the microervices levels associated with the latency we want to predict if we are interested in scaling horizontal resources, or CPU and memory if we are interested in autoscaling vertical resources.

\subsection{End-to-end latency prediction}
\begin{figure*}[!t]
    \centering
    \subfloat[An example of the latency prediction with horizontal resources and the calls at the front-end level in the context of the green trace.]
    {%
        \includegraphics[width=0.4\textwidth]{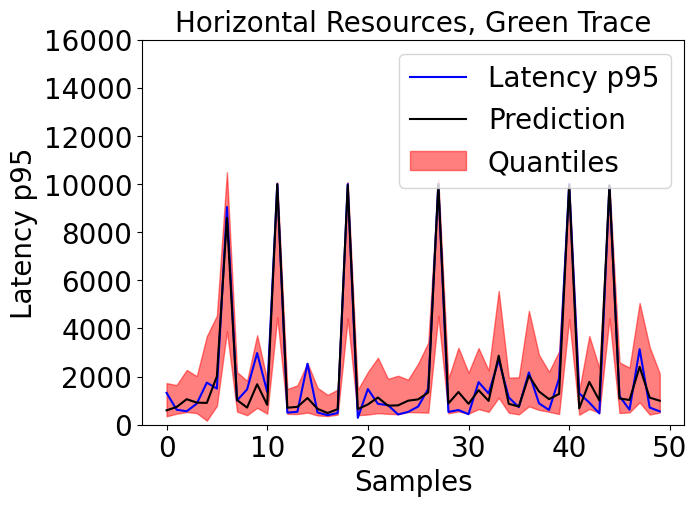}
        \label{fig:pred_horizontal}
    }
    \hspace{0.01\columnwidth} 
    \subfloat[An example of the latency prediction with horizontal resources and the calls at the front-end level in the context of the purple trace.]
    {%
        \includegraphics[width=0.4\textwidth]{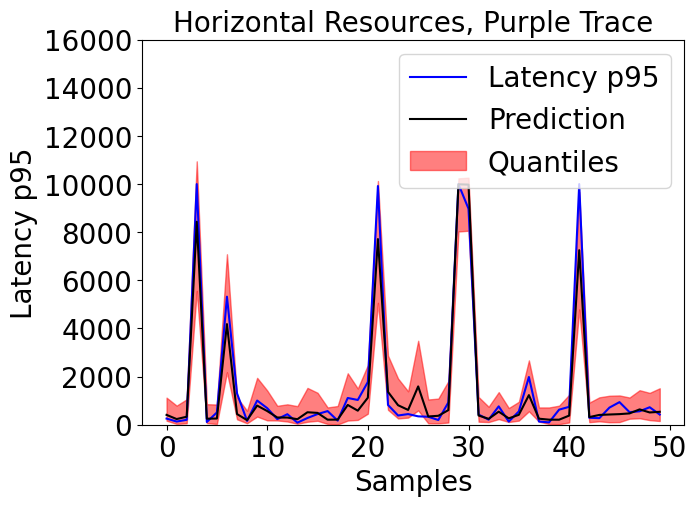}
        \label{fig:pred_vertical}
    }
      \caption{Example of end-to-end latency predictions at the traces level.}
    \label{fig:prediction}   
\end{figure*}

Our experiments do not rely on any autoregressive features to predict the latency. For the TFT model, we include a time index that increments by one for each time step. We standardize each time series separately, ensuring that the values are always positive. To achieve this, we use the EncoderNormalizer, which dynamically scales each encoder sequence during model training. Our model training is conducted using PyTorch Lightning. The distinctive characteristic of the TFT model is its attention mechanism, which attributes different levels of importance to various points in time during latency prediction. This feature provides interpretability to the end-to-end predicted latency. The TFT is designed for multi-horizon prediction, meaning that it can predict future values at multiple time horizons simultaneously. To achieve this, the model incorporates output layers that predict values for each time horizon of interest, allowing it to generate quantile predictions of the end-to-end latency. Additionally, we tune parameters such as a batch size of 32, a learning rate of 0.03, and 20 epochs. Our model architecture includes a hidden size of 8, an attention head size of 1, and a dropout rate of 0.1. Furthermore, we set a maximum prediction length of 50 and a maximum encoder length of 400. We keep these tuning parameters intact across all our experiments. To evaluate the performance of our model, we utilize a quantile loss function. We use early stopping to avoid overfitting and to achieve faster convergence. The variable selection process chooses the relevant data for each time step, encompassing both current and past features and the latency. To handle past metrics, an encoder is employed to incorporate the selected features along with an index indicating their relative time. The encoder processes historical time series data and captures temporal dependencies. It consists of multiple layers of self-attention mechanisms and feedforward neural networks, similar to the encoder in the Transformer model. This encoder encodes the calls at the front-end and the infrastructure metrics into a meaningful representation, which then serves as input to the decoder. Additionally, the decoder takes the features for which latency prediction is desired. In TFT, the decoder primarily generates quantile predictions of the end-to-end latency. Figure \ref{fig:prediction} presents an example of the results of applying the TFT to the presented features and latency. For illustration, we focus on the green and purple traces and the utilization of horizontal resources in the end-to-end latency prediction. In Table \ref{tab:prediction}, we present the application of the TFT to our data and analyze its performance compared to some of the most established regression methods, namely XGBoost, Decision Tree Regressor (DTR), and Random Forest (RF). For this purpose, we use two performance metrics, the Root Mean Square Error (RMSE) and $R^2$. The lowest RMSE and the $R^2$ closer to 1 are associated to the best prediction results. The good performance observed with the green trace in terms of the end-to-end latency prediction is mainly due to the high variability in the number of pods allocated to the microservices belonging to this trace, compared to the other traces, along with the structure of these microservices. This is due to the fact that, in our use case, one of the objectives was to analyze the influence of horizontal resources on the variability of end-to-end latency. Irrespective of the performance metric used, whether the RMSE or $R^2$, there is consistency in terms of the traces that showed good end-to-end latency prediction results. This consistency is also reflected in the features used, whether they are horizontal, vertical, or both.  Table \ref{tab:prediction} demonstrates the advantage of TFT because, in addition to providing good prediction results for the end-to-end latency, the TFT offers interpretability of the prediction results compared to other state-of-the-art regression methods, specifically those presented in this experiment.
Table \ref{tab:resultskrr} presents the application of KRR to the feature importance scores and the desired latency. The results observed in Table \ref{tab:resultskrr} are consistent with those provided in Table \ref{tab:prediction}, where the best performance is obtained when the green trace is considered, and the horizontal resources are used.

\begin{figure*}[!t]
    \centering
    \subfloat[An example of the feature importance associated with the predictions when horizontal resources are used in the context of the green trace.]
    {%
        \includegraphics[width=0.4\textwidth]{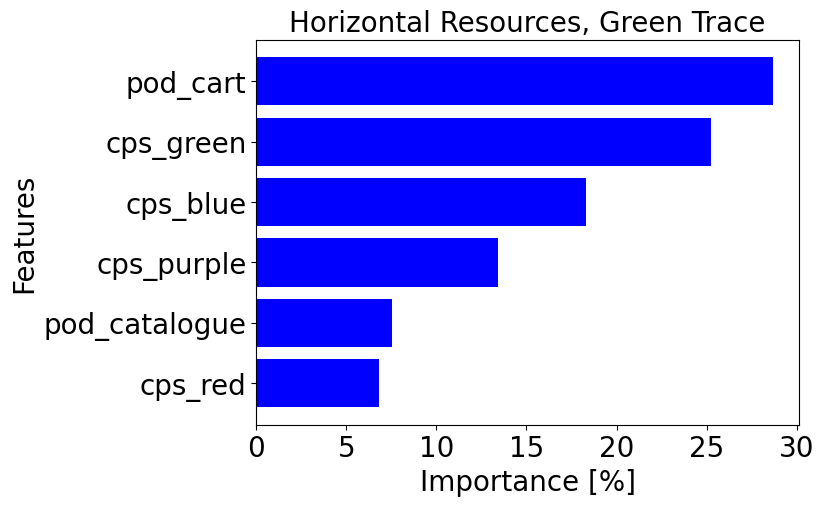}
         \label{fig:feature_importance_horizontal}
    }
    \hspace{0.01\columnwidth} 
    \subfloat[An example of the feature importance associated with the predictions when horizontal resources are used in the context of the purple trace.]
    {%
        \includegraphics[width=0.4\textwidth]{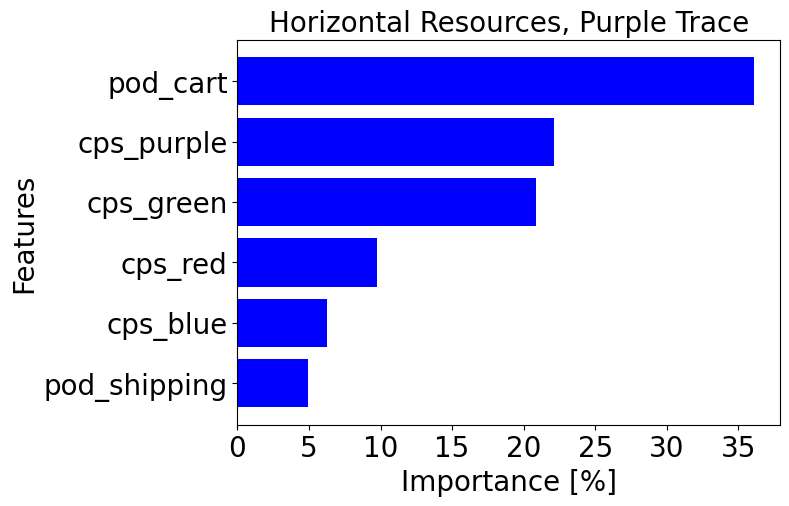}
        \label{fig:feature_importance_vertical}
    }
      \caption{Feature importance associated with the end-to-end latency prediction.}
    \label{fig:feature_importance}   
\end{figure*}

\begin{table*}[ht]
\centering
\resizebox{0.9 \textwidth}{!}{%
\begin{tabular}{@{}lcccccccc@{}} 
                         & \multicolumn{2}{c}{TFT}       & \multicolumn{2}{c}{XGBoost}          & \multicolumn{2}{c}{DTR}        & \multicolumn{2}{c}{RF}    \\ 
\cmidrule(lr){2-3} \cmidrule(lr){4-5}  \cmidrule(lr){6-7} \cmidrule(lr){8-9}
                        & $R^2$ & RMSE             & $R^2$ & RMSE    & $R^2$ & RMSE       & $R^2$ & RMSE     \\ 
\midrule 
horizontal\_resource\_green\_trace     & \textbf{0.971}  & \textbf{496.16}            & 0.92   & 856.23      & 0.90  & 932.31    & 0.91  & 879.12   \\
horizontal\_resource\_purple\_trace        & 0.944  & 672.7328         & 0.89   & 993.12     & 0.87  & 1012.09     & 0.88 & 977.77  \\
vertical\_resource\_green\_trace   & 0.905  & 966.79             & 0.84   & 1020.21     & 0.81  & 1097.36      & 0.82 & 1075.72   \\
vertical\_resource\_purple\_trace         & 0.867  & 1007.96      & 0.79  & 1150.79     & 0.75 & 1297.67      & 0.72  & 1312.19 \\
vertical\_horizontal\_resource\_green\_trace        & 0.957  & 580.0418           & 0.90      & 925.11    & 0.6   & 1485.12    & 0.89  & 912.27    \\
vertical\_horizontal\_resource\_purple\_trace           & 0.81  & 1100.898      & 0.77    & 1300.25  & 0.72   & 1366.32     & 0.75  & 1375.27     
\\ \bottomrule
\end{tabular}%
}
\caption{Performance metrics associated to the end-to-end latency prediction at the front-end level using all the calls and either the horizontal resources, the vertical resources or both. Results obtained with the TFT, XGBoost, Decision Tree Regressor (DTR), and Random Forest (RF).}
\label{tab:prediction}
\end{table*}

\subsection{Feature importance associated with the predictions}
We present in Figure \ref{fig:feature_importance} the feature importance associated with the prediction results shown in Figure \ref{fig:prediction}. One of the primary advantages of the TFT over other DL models is its inherent interpretability, largely attributable to its interpretable multi-head attention mechanisms. With TFT, we can determine the significance of the computing metrics along with the calls at the front-end in the end-to-end latency prediction, a capability present in both the encoder and decoder components. Across various experiments, the consensus is that the number of pods at the cart level is the most influential feature in the latency predictions when the features are composed of horizontal resources. For vertical resources, the memory at the cart level is the most influential. The multi-head attention is crucial for interpretability as it enables the model to focus on different parts of the input data and learn complex temporal dependencies. It also allows the TFT to compute attention weights for different pods and calls at the front-end at various time steps. With these weights, we can interpret which of the number of pods and the calls at the front-end is most relevant for making end-to-end latency predictions at each time step. This provides insights into the relative importance of different features and helps us understand how the model processes and weighs input information when generating the end-to-end latency predictions. Additionally, the multi-head attention enables the TFT to capture both local and global context when making predictions. By using different parts of the input sequence with attention heads, the model can integrate information from nearby and distant time steps to make more informed end-to-end latency predictions. 

\subsection{Autoscaling cloud resources}
We exploit the interpretability results provided by the TFT to implement corrective actions whenever an SLA violation is detected. Our corrective actions either adjust the number of pods (horizontal autoscaling), CPU and memory (vertical autoscaling) or provide insights into how to adapt the characteristics of the calls at the front-end to prevent SLA violations. First, our approach to corrective actions starts with redefining the target variable. For example, if the predicted latency corresponds to an SLA violation of a given percentage, we subtract this percentage from the latency to obtain a new target variable. We then use the feature importance scores provided by the decoder as our new features. The objective becomes estimating the parameters associated with these features and the new target latency. To this end, we use KRR to estimate the parameters required to prevent any SLA violations. For example, for horizontal autoscaling, our KRR will estimate the parameters $\theta_{1}, \cdots, \theta_{6}$ associated with $\theta_{1} f_{1}(cps\_green) + \theta_{2}f_{2}(cps\_blue) +\theta_{3}f_{3}(cps\_purple) + \theta_{4}f_{4}(cps\_red) + \theta_{5}f_{5}(pod\_cart) +\theta_{6}f_{6}(pod\_catalogue) = desired\_latency$, where $cps$ means calls per second. The functions $f_{k}$ are determined by the KRR. The input matrix to the KRR is composed of the feature importance scores provided by the decoder as shown in Figure \ref{fig:feature_importance} and the target variable is the desired latency. We choose a Radial Basis Function (RBF) kernel to capture the relationship between the features and the desired latency. We vary the regularization parameter $\alpha$ and the kernel parameter $\beta$ of the RBF from 0.01 to 10 in steps of 10 and perform a 3-fold cross-validation to determine the best hyperparmeters $\alpha$ and $\beta$ for the KRR. We build six separate KRR models, each corresponding to one feature (e.g., pod\_cart).
We use scipy.optimize to obtain $\theta_{1}, \cdots, \theta_{6}$ that minimize our objective function.

\begin{table}[ht]
\centering
\resizebox{0.9\columnwidth}{!}{%
\begin{tabular}{@{}lcc@{}} 
                         & \multicolumn{2}{c}{KRR}  \\ 
\cmidrule(lr){2-3} 
                        & $R^2$ & RMSE            \\ 
\midrule 
horizontal\_resource\_green\_trace     & \textbf{0.93}  & \textbf{37.23}       \\
horizontal\_resource\_purple\_trace    & 0.89  & 55.34      \\
vertical\_resource\_green\_trace       & 0.86  & 88.33      \\
vertical\_resource\_purple\_trace      & 0.83  & 90.25     \\
vertical\_horizontal\_resource\_green\_trace & 0.91  & 46.41  \\
vertical\_horizontal\_resource\_purple\_trace & 0.77  & 97.34 \\ 
\bottomrule 
\end{tabular}%
}
\caption{Performance obtained after applying the KRR to the score of the feature importance emanating from the multi-head attention of the TFT and the desired end-to-end latency.}
\label{tab:resultskrr}
\end{table}

Table \ref{tab:resultskrr} shows that the best corrective action is obtained for the green trace using the horizontal resources, which is congruent with the results obtained for the end-to-end latency prediction using the green trace and the horizontal resources.

\section{Roadmap to deployment}
\label{roadmap}
This work represents the development of the AI component of our multi-cloud manager system. It needs to be integrated into the overall multi-cloud manager system, where our developed approach will operate in conjunction with several other components. These include the monitoring system that collects data from the microservices, the ontology and semantic system that provides a formal representation of the data for convenient access and analysis, the resource exposure and discovery component that indicates the status of resources across the multi-cloud environment to decide where to run the microservices-based architecture, and the network programmability component that will effectively enforce the autoscaling results presented in this work. Following the integration of the AI component into the multi-cloud manager system, further evaluations in various settings are required to enable the effective and large-scale deployment of our framework. This is necessary because, although the approach provides good results that can be used to adjust the actionable features, the experimentation of the approach in diverse environments and the quantification of the reduction in SLA violations resulting from the framework need to be assessed before an extensive rollout.

\section{Conclusions}
\label{conclusions}
This work establishes a foundation for the efficient autoscaling of cloud resources in microservices-based applications. To achieve this, we developed an innovative approach consisting of three key steps. The first step provides an interpretable end-to-end latency prediction, which enables the detection of potential SLA violations. In the case of an SLA violation, we utilize the interpretability results from the multi-head attention of the TFT, combined with the KRR, to identify the parameters that need adjustment and the extent of the adjustments required to correct the SLA violations. Following this step, we implement the autoscaling process. The performance metrics of our results demonstrate the effectiveness of our approach and its practical merit. This work is the first to use the interpretability of the Transformer to build autoscalers and makes a significant contribution to cloud providers by enhancing their ability to maintain SLA compliance efficiently through dynamic resource scaling.

\bibliographystyle{plainnat} 
\bibliography{references}

\end{document}